\documentclass[opre,nonblindrev]{informs3}

\DoubleSpacedXI 

\usepackage{endnotes}
\let\footnote=\endnote

\usepackage{natbib}
 \bibpunct[, ]{(}{)}{,}{a}{}{,}%
 \def\bibfont{\small}%
\TheoremsNumberedThrough     
\ECRepeatTheorems

\EquationsNumberedThrough    
\usepackage{hyperref}
\usepackage[ruled,vlined]{algorithm2e}
\usepackage{algorithmic}

\def\C{\mathcal{C}}
\def\A{\mathcal{A}}
\def\EE{\mathcal{E}}
\def\S{\mathcal{S}}

\def\X{\mathcal{X}}

\def\E{\mathbb{E}}
\def\P{\mathbb{P}}
\def\R{\mathbb{R}}

\def\1{\mathbf{1}}

\def\maketitle{%
  \newpage
  \theARTICLETITLE
  \theARTICLEAUTHORS
  \theARTICLEABSTRACT
  \theARTICLERULE
}

\def\setoddRH{\hbox to \textwidth{\fs.7.8.\tabcolsep0pt
  \begin{tabular*}{\textwidth}[b]{l@{\extracolsep\fill}r}
   {\theRRHFirstLine}&\raisebox{0pt}[0pt][0pt]{\fs.10.10.\thepage}\\[-4pt]
  \rlap{\VRHDW{0.5pt}{0pt}{\textwidth}}&\\
  \end{tabular*}}}
\def\setevenRH{\hbox to \textwidth{\fs.7.8.\tabcolsep0pt
  \begin{tabular*}{\textwidth}[b]{l@{\extracolsep\fill}r} {\theRRHFirstLine}&\raisebox{0pt}[0pt][0pt]{\fs.10.10.\thepage}\\[-4pt]
  \rlap{\VRHDW{0.5pt}{0pt}{\textwidth}}&\\
  \end{tabular*}}}
  
\begin{document}


\RUNAUTHOR{Kazerouni and Wein}

\RUNTITLE{Best Arm Identification in Generalized Linear Bandits}

\TITLE{Best Arm Identification in Generalized Linear Bandits}

\ARTICLEAUTHORS{%
\AUTHOR{Abbas Kazerouni}
\AFF{Department of Electrical Engineering, Stanford University, Stanford, CA 94305, \EMAIL{abbask@stanford.edu}} 
\AUTHOR{Lawrence M. Wein}
\AFF{Graduate School of Business, Stanford University, Stanford, CA 94305, \EMAIL{lwein@stanford.edu}}
} 

\ABSTRACT{%
Motivated by drug design, we consider the best-arm identification problem in generalized linear bandits. More specifically, we assume each arm has a vector of covariates, there is an unknown vector of parameters that is common across the arms, and a generalized linear model captures the dependence of rewards on the covariate and parameter vectors. The problem is to minimize the number of arm pulls required to identify an arm that is sufficiently close to optimal with a sufficiently high probability. Building on recent progress in best-arm identification for linear bandits \citep{xu2018}, we propose the first algorithm  for best-arm identification for generalized linear bandits, provide theoretical guarantees on its accuracy and sampling efficiency, and evaluate its performance in various scenarios via simulation.
}%


\KEYWORDS{best arm identification, generalized linear bandits, sequential clinical trial} 

\maketitle

%


\section{Introduction}
\label{sec::intro}
The multi-armed bandit problem is a prototypical model for optimizing the tradeoff between exploration and exploitation. We consider a pure-exploration version of the bandit problem known as the best-arm identification problem, where the goal is to minimize the number of arm pulls required to select an arm that is - with sufficiently high probability -- sufficiently close to the best arm. We assume that each arm has an observable vector of covariates or features, and there is an unknown vector of parameters (of the same dimension as the vector of features) that is common across arms. Whereas in a linear bandit the mean reward of an arm is the linear predictor (i.e., the inner product of the parameter vector and the feature vector), in our generalized linear model the mean reward is related to the linear predictor via a link function, which allows for mean rewards that are nonlinear in the linear predictor, as well as binary or integer rewards (via, e.g., logistic or Poisson  regression). Hence, with every pull of an arm, the decision maker refines the estimate of the unknown parameter vector and learns simultaneously about all arms. 

Our motivation for studying this version of the bandit problem comes from drug design. The field of drug design is immense \citep{drug}. There are different types of drugs (e.g., organic small molecules or biopolymer-based, i.e. biopharmaceutical, drugs), several phases of development, and a variety of experimental and computational methods used to select compounds with favorable properties. Three characteristics of our bandit problem make it ideally suited to certain subproblems within the drug design process. The first characteristic is the desire to run as few experiments as possible with the goal of selecting a particular compound (i.e., an arm in our model) for the next stage of analysis or testing. Hence, the actual performance of the selected arms during testing is of secondary importance (these pre-clinical tests do not involve human subjects), giving rise to a pure exploration problem. Second, drugs can often be described by a vector of features, which may include calculated properties of molecules, or the three-dimensional structure of the molecule and how it relates to the three-dimensional structure of the biological target. Finally, in many of these experimental settings, a generalized linear model provides a better fit than the linear model. This is true when the outcome is binary; e.g., in animal experiments, there is either survival or death, or either the presence or absence of disease, and some {\it in vitro} assays are qualitative (i.e., binary output). It is also true when the outcome is counting data (e.g., the number of cells or animals that survived or died), or when the relationship between the outcome and the best linear predictor is nonlinear. 

There may also be other potential applications of this model, e.g., where the aim is to choose the best (non-personalized) ad for an advertising campaign, and the observed output is binary (e.g., the ad led to a purchase or a click-through). 

\noindent {\bf Literature Review.} A recent review \citep{6} categorizes bandit problems into three types: Markovian (maximizing discounted rewards, often in a Bayesian setting, and exemplified by the celebrated result in \cite{gittins}), stochastic (typically minimizing regret in a frequentist setting, with independent and identically distributed  rewards \citep{lai}), and adversarial (a worst-case setting, where an adversary chooses the rewards \citep{auer2002b}). We study the stochastic problem, but consider best-arm identification rather than regret minimization. In passing, we note that the ranking-and-selection problem \citep{kim} in the simulation literature has a similar goal to best-arm identification, although evaluating an alternative requires running simulation experiments. 

There is a vast literature on best-arm identification in the multi-arm bandit setting with independent arms, where pulling one arm does not reveal any information about the reward of other arms \citep{evandar2006, bubeck2009, audibert2010, gabillon2012, kaly2012, karnin2013, chen2014}. Different algorithms have been developed for variants of this setting, most of which use gap-based exploration where arms are played to reduce the uncertainty about the gaps between the rewards of pairs of arms. Because playing an arm reveals information only about that arm, each arm needs to be played several times to reduce the uncertainty about its reward. As such, these algorithms can be practically implemented only when the number of available arms is relatively small.

Rather than assume independent arms, our formulation considers parametric arms, where each arm has a covariate vector and there is an unknown parameter vector that is common across arms. There has been considerable work on the linear parametric bandits (i.e., the mean reward of an arm is the inner product of its covariate vector and the parameter vector) under the minimum-regret objective (e.g., \cite{4in20,rus2010} and references therein) as well as alternative probabilistic models of arm dependence (e.g., \cite{28} and references therein). In addition, contextual bandit models allow additional side information in each round, which can model, e.g., patient information in clinical trials or consumer information in online advertising (e.g., \cite{wang2005, seldin2011, zeevi}). Relevant for our purposes, the generalized linear parametric bandit, which uses an inverse link function to relate the linear predictor and the mean reward, has been studied under regret minimization \citep{filippi2010, li2017}. 

However, relatively little work exists on best-arm identification in parametric bandits. The first analysis of best-arm identification in linear bandits \citep{soare2014} proposed a static exploration algorithm that was inspired by the transductive experiment design principle \citep{yu2006}. This algorithm determines the sequence of to-be-played actions before making any observations and fails to adapt to the observed rewards. 
Recently \citep{xu2018}, major progress has been made via the first adaptive algorithm for best-arm identification in linear bandits. These authors  design a gap-based exploration algorithm by employing the standard confidence sets constructed in the literature for linear bandits under regret minimization. 

\noindent{\bf Our Contribution.} We adapt the gap-based exploration algorithm of \cite{xu2018} from the linear setting to the generalized linear case, which requires us to derive confidence sets for reward gaps between different pairs of arms. In the regret-minimization setting, the typical approach to the linear bandit is to develop a confidence set for the unknown parameter vector that governs the rewards of all arms, whereas in the best-arm setting, a confidence set on the reward gaps is needed. In the best-arm identification for the linear bandit \citep{xu2018}, the authors were able to convert the confidence set for the parameter vector into efficient confidence sets for the reward gaps. However, this approach breaks down in the generalized linear bandit; i.e., naively converting the confidence set for the parameter vector in \cite{filippi2010} into confidence sets for reward gaps between arms leads to extremely loose confidence sets, which strongly degrades the performance of the gap-based exploration algorithm. Rather than use this indirect method, we build gap confidence sets directly from the data. 

The remainder of this paper is organized as follows. In Section \ref{sec::formulation}, we formulate the best-arm identification problem for generalized linear bandits. We describe our algorithm in Section \ref{sec::alg} and establish theoretical guarantees in Section \ref{sec::theo}. We provide simulation results in Section \ref{sec::sim} and offer concluding remarks in Section \ref{sec::concl}.

\section{Problem Formulation}
\label{sec::formulation}
Consider a decision maker who is seeking to find the best among a set of $K$ available arms. We let $[K]=\{1,2,\cdots,K\}$ denote the set of possible arms. There is a feature vector $x^a\in\R^d$ associated with arm $a$, for $a\in[K]$. These feature vectors are known to the decision maker and each summarizes the available information about the corresponding arm. We employ a generalized linear model \citep{glmbook} and assume that the reward of each arm $a$ has a particular distribution in an exponential family with mean 
\begin{equation}
\label{reward1}
\mu_a = \mu(\theta^\intercal x^a),
\end{equation}
where $\theta \in\R^d$ is an unknown parameter that governs the reward of all arms and $\mu:\R\to \R$ is a strictly increasing function known as the inverse link function. Different choices for the function $\mu(z)$ in \eqref{reward1} result in modeling different reward structures inside the exponential family. For example, choosing $\mu(z) = e^z$ and $\mu(z) = 1/(1+e^{-z})$ correspond to a Poisson regression model and a logistic regression model, respectively. 

The decision maker chooses an arm to play in each round $t$. If arm $a_t$ is played in round $t$, a stochastic reward $r_t$ is observed, which satisfies  
$$\E[r_t|a_t] = \mu_{a_t} = \mu(\theta^\intercal x^{a_t}).$$

Let $a^*=\argmax_{a\in[K]} ~\mu_a$ be the optimal arm; i.e., the arm with the highest expected reward. By exploring different arms, the decision maker is trying to find the optimal arm as soon as possible based on the noisy observations. Let $\tau$ be a stopping time that dictates whether enough evidence has been gathered to declare the optimal arm. The declared optimal arm is denoted by $\hat a_{\tau}$. An exploration strategy can be represented by $\S=(\A,\tau)$, where, at any time $t$, $\A$ is a function mapping from the previous observations $\{(a_i,r_i)\}_{i=1}^{t-1}$ to the arm $a_t$ to be played next, and $\tau$ determines whether enough information has been gathered to  declare the optimal arm, $\hat a_{\tau}$. Because finding the exact optimal arm may require a prohibitively large amount of exploration, the performance of an exploration strategy is evaluated via the following relaxed criterion. 
\begin{definition}
\label{ep-del-def}
Given $\epsilon>0$ and $\delta\in (0,1)$, an exploration strategy $\S=(\A,\tau)$ is said to be $(\epsilon,\delta)-$optimal if 
\begin{equation}
\label{bestcosntraint}
\P\left[\mu(\theta^\intercal x^{a^*}) -\mu(\theta^\intercal x^{\hat a_\tau})\geq \epsilon\right]\leq \delta.
\end{equation} 
\end{definition}
In this definition, $\epsilon$ denotes an acceptable region around the optimal arm and $1-\delta$ represents the confidence in identifying an arm within this region.  This criterion  relaxes the notion of optimality by allowing the exploration strategy to return a sufficiently good -- but not necessarily optimal -- arm. With this definition in place, the decision maker's goal is to design an $(\epsilon,\delta)-$optimal exploration strategy with the smallest possible stopping time. 

Before proceeding to the algorithm, we introduce additional notation and state a set of regularity assumptions. We let $\X=\{x^a\}_{a\in[K]}$ denote the set of feature vectors  and assume that feature vectors, the unknown reward parameter and the rewards are bounded; i.e., there exist $S,L,R>0$ such that $\|\theta\|_2\leq S$, $\|x\|_2\leq L ~\forall x\in \X$, and $r_t\leq R$ almost surely for all $t$. We also assume that $\mu$ is continuously differentiable, Lipschitz continuous with constant $k_\mu$ and satisfies $c_\mu = \inf_{\theta:\|\theta\|\leq S,~x\in \X} \dot{\mu}(\theta^\intercal x)>0$. For example, in the case of logistic regression, $R = 1, k_\mu = 1/4$ and $c_\mu$ depends on $\sup_{\theta:\|\theta\|\leq S,~x\in \X} |\theta^\intercal x|$. We define the gap between any two arms $i,j\in[K]$ to be $\Delta(i,j) = \mu_i - \mu_j = \mu(\theta^\top x^i)- \mu(\theta^\top x^j)$ and define the optimal gap associated to an arm $i\in [K]$ as 
\begin{equation}
\label{optgapdef}
\Delta_i = \left\{\begin{aligned}
&\mu(\theta^\top x^{a^*})- \mu(\theta^\top x^i) &&\mbox{if }i\neq a^*\\
&\mu(\theta^\top x^i) - \max_{j\neq i} \mu(\theta^\top x^j) &&\mbox{if }i= a^* .
\end{aligned} \right.
\end{equation}
Finally, for any positive semi-definite matrix $A$, we let $\|x\|_A = \sqrt{x^\top A x}$.

\section{The Proposed Algorithm}
\label{sec::alg}

In this section, we propose an exploration strategy for the problem formulated in Section~\ref{sec::formulation}. Following \cite{xu2018}, our algorithm consists of the following steps:
\begin{enumerate}
\item Build confidence sets for the pairwise gaps between arms,
\item Identify the potential best arm and an alternative arm that has the most ambiguous gap with the best arm,
\item Play an arm to reduce this ambiguity.
\end{enumerate}
These steps are repeated sequentially until the ambiguity in step 2 drops below a certain threshold. 

The confidence sets are derived in Subsection~\ref{subsec:conf} and the algorithm is presented in Subsection~\ref{subsec:alg}. 

\subsection{Confidence Sets for Gaps}
\label{subsec:conf}
To build the confidence sets for reward gaps, we follow ideas in \cite{filippi2010} but  develop confidence sets directly for gaps instead of arm rewards.

Let $(a_1,r_1),(a_2,r_2),\cdots,(a_{t-1},r_{t-1})$ be the history of  actions played and random  rewards observed prior to period $t$, and let $x_l = x^{a_l}$ be shorthand notation for the feature vector associated with the arm played in period $l$. For any $t>0$, let $M_t = \sum_{l=1}^{t-1} x_l x_l^\top$ be the empirical covariance matrix and assume it is nonsingular for any $t>E$ for some fixed value $E>d$. We let $\lambda_0>0$ be the minimum eigenvalue of $M_{E+1}$ and define $\kappa=\sqrt{3+2\log(1+2L^2/\lambda_0)}$. Given the observations by the start of period $t$, the Maximum Likelihood (ML) estimate of the reward parameter, $\theta_t$, solves the equation
 \begin{equation}
\label{mlequation}
\sum_{l=1}^{t-1} (r_l - \mu(\theta_t^\intercal x_l)) x_l = 0.
\end{equation}
Based on the estimated reward parameter, we can take 
\begin{equation}
\label{gapest}
\Delta_t(i,j) = \mu(\theta_t^\top x^i) - \mu(\theta_t^\top x^j)
\end{equation}
as an estimate for the gap between arms $i, j\in[K]$, which is a function of  the observations made prior to period $t$.  

Given $\delta\in(0,1)$, let
\begin{equation}
\label{confwidth}
C_t = \alpha\sqrt{2d\log t \log\Bigl(\frac{\pi^2dt^2}{6\delta}\Bigr)},
\end{equation}
where $\alpha$ is a tunable parameter and $C_t$ is a time-varying quantity that scales the width of the confidence sets for all pairs of arms. We set 
\begin{equation}
\label{alpha}
\alpha = \frac{2\kappa R}{c_{\mu}}
\end{equation}
in the theoretical analysis in Section~\ref{sec::theo}, and set $\alpha$ so as to achieve robust performance across a variety of scenarios in the computational study in Section~\ref{sec::sim}. We consider the following confidence set for the gap between any two arms $i,j\in[K]$ based on the observations made prior to period $t$:
\begin{equation}
\label{gapconfdef}
\C_t(i,j) = \left\{\Delta\in\R:|\Delta-\Delta_t(i,j)|\leq C_t \max_{c,c'\in[c_\mu,k_\mu]}\|cx^i-c'x^j\|_{M_t^{-1}}\right\}.
\end{equation}
The confidence set defined in \eqref{gapconfdef} is centered around the estimated gap between the two arms in~(\ref{gapest}) and, as we will show in Section \ref{sec::theo}, contains the true gap with high probability. To further simplify the representation, we let 
\begin{equation}
\label{betadef}
\beta_t(i,j) = C_t \max_{c,c'\in[c_\mu,k_\mu]}\|cx^i-c'x^j\|_{M_t^{-1}}
\end{equation}
represent the width of the confidence set in \eqref{gapconfdef}. 

Although we follow the basic approach in \cite{filippi2010} in deriving these confidence sets, it is worth noting that they cannot be deduced from the results presented in \cite{filippi2010}. More precisely, an immediate use of the results in that paper gives rise to a very loose confidence set that is similar to \eqref{gapconfdef} except that  $\max_{c,c'\in[c_\mu,k_\mu]}\|cx^i-c'x^j\|_{M_t^{-1}}$ is replaced by the looser factor $\left(\|x^i\|_{M_t^{-1}} +\|x^j\|_{M_t^{-1}} \right)$.
 
\TableSpaced
\begin{minipage}[t]{0.5\textwidth}
\begin{algorithm}[H]
\caption{GLGapE }
   \label{alg1}
   \KwIn {$\X,\epsilon, \delta, E,\alpha, S,L,R$}
   \KwOut {approximately best arm $\hat a_\tau$}
\begin{algorithmic}[1]
    \STATE{\bfseries Initial Exploratory Phase:} Play $E$ random arms and gather observations in $D_E$, and initialize $T_a(t)$'s for the played arms
    
   \FOR{$t>E$}
   \STATE \texttt{//select a gap to examine}
   \STATE $i_t,j_t,B(t),y_t\leftarrow \mbox{select-gap}(\X,D_{t-1})$
   \IF{$B(t)\leq \epsilon$}
   \STATE {\bf return} $i_t$ as the best arm
   \ENDIF
   \STATE \texttt{//select  an action}
   \STATE $a_t\leftarrow$select-arm$(\X,D_{t-1},T(t),y_t)$
   \STATE Play $a_t$
   \STATE Observe $r_t$
   	\STATE $D_t= D_{t-1}\cup\{(x_t,r_t)\}$
   	\STATE $T_a(t)=T_a(t-1)+ 1$
   \ENDFOR
  
  \end{algorithmic}
  \end{algorithm}
\end{minipage}
\begin{minipage}[t]{0.5\textwidth}
\begin{algorithm}[H]
   \caption{select-gap}
   \label{alg2}
   \KwIn {$\X,D$}
   \KwOut {a gap to be examined}
\begin{algorithmic}[1]
	\STATE Find the ML estimate $\theta_t$
	\STATE $i_t\leftarrow \argmax_{i\in [K]}~\mu(\theta_t^\top x^i)$
	\STATE $j_t\leftarrow \argmax_{j\in[K], j\neq i_t}~\Delta_t(j,i_t) + \beta_t(i_t,j)$
	\STATE Find $c_1,c_2 = \argmax_{c,c'\in[c_\mu,k_\mu]}\|cx^{i_t} - c'x^{j_t}\|_{M_t^{-1}}$
	\STATE Let $y_t = c_1x^{i_t} - c_2x^{j_t}$
	\STATE $B(t)\leftarrow \Delta_t(j_t,i_t) + \beta_t(i_t,j_t)$
	\STATE {\bf return} $i_t,j_t,B(t),y_t$
\end{algorithmic}
\end{algorithm} 

\begin{algorithm}[H]
   \caption{select-arm}
   \label{alg3}
   \KwIn {$\X,D,T,y$}
   \KwOut {arm to be played}
\begin{algorithmic}[1]
	\STATE Find $w^*_1,\cdots,w^*_K$ as the solution of \\ $\argmin_{\{w_a\}}~\sum_{a=1}^K |w_a|~~~\mbox{s.t.  }~~\sum_{a=1}^K w_a x^a = y$
	\STATE Determine $p_a =\frac{|w_a^*|}{\sum_{k=1}^K |w_k^*|}$ for $a=1,\cdots,K$ 
	\STATE Find $\hat a_t = \argmin_{a\in[K]:p_a>0}~\frac{T_a(t)}{p_a}$
	\STATE {\bf return} $\hat a_t$
\end{algorithmic}
\end{algorithm} 
\end{minipage}
\DoubleSpacedXI

\subsection{The Algorithm}
\label{subsec:alg}
With the confidence sets established, we are now ready to describe our proposed algorithm. Details and successive steps of the proposed algorithm are presented in Algorithms \ref{alg1}-\ref{alg3}. Following the gap-based exploration scheme, the proposed algorithm consists of two major components that are described below. 

\begin{description}
\item[\bf{Selecting a Gap to Explore:}]
The algorithm starts by playing $E>d$ random arms such that the empirical covariance matrix $M_{E+1}$ is nonsingular. At any subsequent period $t>E$, the algorithm first finds the empirically best arm $i_t = \argmax_{i\in[K]} \mu(\theta_t^\top x^i)$. Then, to check whether  this arm is within $\epsilon$ distance of the true optimal arm, the algorithm takes a pessimistic approach. In particular, it finds another arm that is the most advantageous over arm $i_t$ within the gap confidence sets; i.e., $j_t = \argmax_{j\in[K], j\neq i_t}\Delta_t(j,i_t) + \beta_t(j,i_t)$. Note that this pessimistic gap consists of two components: the estimated gap and the uncertainty in the gap. 

If this pessimistic gap is less than $\epsilon$, the algorithm stops and declares the empirically best arm as the optimal arm. Otherwise, it selects an arm to reduce the uncertainty component in the identified gap.  According to \eqref{betadef}, this uncertainty is governed by $\|y_t\|_{M_t^{-1}}$ where $y_t = c_1x^{i_t} - c_2x^{j_t}$ and $c_1,c_2 = \argmax_{c,c'\in[c_\mu,k_\mu]}\|cx^{i_t} - c'x^{j_t}\|_{M_t^{-1}}$. 

\item[\bf{Selecting an Arm:}]
While there are different ways to reduce $\|y_t\|_{M_t^{-1}}$, we follow the approach in  \cite{xu2018}. With a slight abuse of notation, let us define $M_{\mathbf{z}_n} = \sum_{i=1}^n z_iz_i^\top$ for any sequence of feature vectors $\mathbf{z}_n = (z_1,z_2,\ldots,z_n)\in\X^n$, where $n$ represents a generic time period. Let $\mathbf{z}_n^*(y) = \argmin_{\mathbf{z}_n\in\X^n} \|y\|_{M^{-1}_{\mathbf{z}_n}}$ be the sequence of feature vectors that would have minimized the uncertainty in the direction of $y$. For each arm $a\in[K]$, let $p_a(y)$ denote the relative frequency of $x^a$ appearing in the sequence $\mathbf{z}_n$ when $n\to\infty$. As has been shown in Section 5.1 of \cite{xu2018}, 
\begin{equation}
\label{prel}
p_a(y) = \frac{|w^*_a(y)|}{\sum_{k=1}^K |w_k^*(y)|},
\end{equation}
where $w^*(y)\in\R^K$ is the solution of the linear program
\begin{equation}
\label{linprog}
\min_{w\in\R^K}~\|w\|_1~~~~~\mbox{s.t. }~~~\sum_{k=1}^K w_k x^k = y.
\end{equation}
To minimize the uncertainty in the direction of $y_t$, the algorithm plays the arm
\begin{equation}
\label{playedaction}
a_{t+1} = \argmin_{a\in[K]:p_a(y_t)>0} ~\frac{T_a(t)}{p_a(y_t)},
\end{equation}
where $T_a(t)$ is the number of times arm $a$ has been played prior to period $t$. 

\end{description}

\section{Theoretical Analysis}
\label{sec::theo}
In this section, we provide theoretical guarantees for the performance of the proposed algorithm. We prove that the algorithm indeed finds an $(\epsilon,\delta)-$optimal arm in Subsection \ref{ssec::theo:optimal} and provide an upper bound on the stopping time of the algorithm in Subsection~\ref{ssec:theo:tau}. 

\subsection{$(\epsilon,\delta$)-Optimality of the Proposed Algorithm}
\label{ssec::theo:optimal}
We start by proving that the confidence sets constructed in Section~\ref{subsec:conf} hold with high probability at all times. 
\begin{proposition}
\label{prop0}
Fix $\delta$ and $t$ such that $0<\delta<\min(1,d/e)$ and $t>\max(2,d)$, where $d$ is the feature dimension. Then the following holds with probability at least $1-\delta$:
\begin{equation}
\label{conf1}
\forall x,x'\in\X:~~|\left[\mu(\theta_t^\intercal x) - \mu(\theta_t^\intercal x')\right]-\left[\mu(\theta^\intercal x) - \mu(\theta^\intercal x')\right]|\leq D_t(\delta) \max_{c,c'\in [c_\mu,k_\mu]}\|c x - c' x'\|_{M_t^{-1}},
\end{equation}
where 
\begin{equation}
\label{Dt}
D_t(\delta)  = \frac{2\kappa R}{c_\mu}\sqrt{2d\log t \log(d/\delta)}.
\end{equation}
\end{proposition}
The proof is a slight modification of the proof of Proposition 1 in \cite{filippi2010}. 
\proof{Proof.}
Let $g_t(\beta) = \sum_{l=1}^{t-1} \mu(\beta^\top x_l)x_l$. According to \eqref{mlequation}, the ML estimate $\theta_t$ satisfies $g_t(\theta_t) = 0$. By the mean value theorem,   there exist points $z,z'$ with $c = \dot{\mu}(z),c' = \dot{\mu}(z')$ such that 
\begin{align}
|\left[\mu(\theta_t^\intercal x) - \mu(\theta_t^\intercal x')\right]-\left[\mu(\theta^\intercal x) - \mu(\theta^\intercal x')\right]| & = |\left[\mu(\theta_t^\intercal x) - \mu(\theta^\intercal x)\right]-\left[\mu(\theta_t^\intercal x') - \mu(\theta^\intercal x')\right]|,\nonumber\\
& = \left|\left[\dot{\mu}(z)(\theta_t - \theta)^\top x - \dot{\mu}(z')(\theta_t - \theta)^\top x' \right]\right|, \nonumber\\
& = \left|(\theta_t - \theta)^\top (c x - c'x')\right|. \label{linear1}
\end{align}

On the other hand, by the fundamental theorem of calculus, we have 
\begin{equation}
\label{gt}
g_t(\theta_t) - g_t(\theta) = G_t(\theta_t - \theta),
\end{equation}
where 
$$G_t = \int_0^1 \nabla g_t(s\theta + (1-s)\theta_t)~ds.$$
The definition of $g_t(\beta)$ implies that for any $\beta$, 
$$\nabla g_t(\beta) = \sum_{l=1}^{t-1} x_l x_l^\top \dot{\mu}(\beta ^\top x_l).$$
It follows that $G_t\succeq c_\mu M_t \succeq c_\mu M_d \succeq \lambda_0 I\succ 0$. Hence, $G_t$ and $G_t^{-1}$ are positive definite and nonsingular. Therefore, from \eqref{linear1} and \eqref{gt}, it follows that 
\begin{align}
\left|\left[\mu(\theta_t^\intercal x) - \mu(\theta_t^\intercal x')\right]-\left[\mu(\theta^\intercal x) - \mu(\theta^\intercal x')\right]\right| & = \left|(c x - c'x')^\top G_t^{-1}\left[g_t(\theta_t) - g_t(\theta) \right]\right|,\nonumber \\
&\leq \|cx - c'x'\|_{G_t^{-1}} \|g_t(\theta_t) - g_t(\theta)\|_{G_t^{-1}}.\label{eq15a}
\end{align}

The inequality $G_t\succeq c_\mu$ implies that $G_t^{-1}\preceq c_\mu^{-1}M_t^{-1}$, and hence $\|y\|_{G_t^{-1}}\leq \frac{1}{\sqrt{c_\mu}} \|y\|_{M_t^{-1}}$ for arbitrary $y\in\R^d$. Thus, by~(\ref{eq15a}) we get
\begin{equation}
\label{bound1} 
\left|\left[\mu(\theta_t^\intercal x) - \mu(\theta_t^\intercal x')\right]-\left[\mu(\theta^\intercal x) - \mu(\theta^\intercal x')\right]\right| \leq \frac{2}{c_\mu} \|cx - c'x'\|_{M_t^{-1}} \|g_t(\theta_t) - g_t(\theta)\|_{M_t^{-1}}.
\end{equation}
As has been shown in the proof of Proposition 1 in \cite{filippi2010}, 
\begin{equation}
\label{gtbound}
 \|g_t(\theta_t) - g_t(\theta)\|_{M_t^{-1}} \leq \kappa R\sqrt{2d\log t\log(d/\delta)}
\end{equation}
holds with probability at least $1-\delta$. Combining \eqref{bound1} and \eqref{gtbound} shows that 
\begin{equation}
\label{eq17a}
\left|\left[\mu(\theta_t^\intercal x) - \mu(\theta_t^\intercal x')\right]-\left[\mu(\theta^\intercal x) - \mu(\theta^\intercal x')\right]\right|\leq D_t(\delta) \|cx - c'x'\|_{M_t^{-1}}
\end{equation}
holds with probability at least $1-\delta$. Finally, because $c,c'\in[c_\mu,k_\mu]$, taking the maximum over $c,c'$ on the right side of~(\ref{eq17a}) completes the proof. \Halmos 
\endproof

The following theorem, which is a direct consequence of Proposition~\ref{prop0}, shows that the confidence sets in~(\ref{gapconfdef}) contain the true gaps at all times with high probability. We define event $\EE$ to be
\begin{equation}
\label{eventE}
\mathcal E  =\left\{\forall~t>\max(2,d),\forall i,j\in[K]: \Delta(i,j)\in\C_t(i,j)\right\},
\end{equation}
and throughout this section we set $\alpha$ according to~(\ref{alpha}).

\begin{theorem}
\label{prop1}
Let $\delta$ be such that $0<\delta<\min(1,d/e)$. Then event $\mathcal{E}$ occurs with probability at least $1-\delta$.
\end{theorem}
\proof{Proof. }
For any $t>\max(2,d)$, let $\delta_t=\frac{6\delta}{\pi^2t^2}$ and define the event $\mathcal{E}_t$ as
$$\mathcal{E}_t=\left\{\forall x,x'\in\X:~~|\left[\mu(\theta_t^\intercal x) - \mu(\theta_t^\intercal x')\right]-\left[\mu(\theta^\intercal x) - \mu(\theta^\intercal x')\right]|\leq D_t(\delta_t) \max_{c,c'\in [c_\mu,k_\mu]}\|c x - c' x'\|_{M_t^{-1}}\right\}.$$
Applying Proposition \ref{prop0} for $\delta_t$ implies that $\P[\mathcal{E}_t] \geq 1-\delta_t$. The union bound then gives
$$\P[\cap_{t=1}^\infty\mathcal{E}_t] \geq 1-\sum_{t=1}^\infty \delta_t = 1-\frac{6\delta}{\pi^2}\sum_{t=1}^\infty \frac{1}{t^2} = 1-\delta.$$
The proof is completed by noting that $\mathcal{E} = \cap_{t=1}^\infty\mathcal{E}_t$.\Halmos
\endproof

With the confidence sets established, the next theorem proves $(\epsilon,\delta)-$optimality of the proposed algorithm. 
\begin{theorem}
\label{optthm}
Let $\epsilon>0$ and $0<\delta<\min(1,d/e)$ be arbitrary.  Then the proposed algorithm is  $(\epsilon,\delta)-$optimal; i.e., at its stopping time, the algorithm returns an arm $\hat a_\tau$ such that
$$\P[\Delta(a^*,\hat a_\tau)>\epsilon]<\delta.$$
\end{theorem}
\proof{Proof.}
Let $\tau$ be the stopping time of the algorithm and let $\hat a_\tau$ be the returned arm. Suppose that $\Delta(a^*,\hat a_\tau)> \epsilon$. Then, according to line 5 of Algorithm \ref{alg1}, we have
$$\Delta(a^*,\hat a_\tau) >\epsilon\geq B(\tau)\geq \Delta_\tau(a^*,\hat a_\tau) + \beta_\tau(a^*,\hat a_\tau).$$
From this, we get that
$$\Delta(a^*,\hat a_\tau) - \Delta_\tau(a^*,\hat a_\tau)>\beta_\tau(a^*,\hat a_\tau),$$
which means that event $\EE$ does not occur. According to Theorem~\ref{prop1}, this can happen with probability at most $\delta$. Thus, $\Delta(a^*,\hat a_\tau)> \epsilon$ happens with probability at most $\delta$.  \Halmos
\endproof

\subsection{An Upper Bound on the Stopping Time}
\label{ssec:theo:tau}
In this subsection, we study the sample complexity of the proposed algorithm. The following theorem provides an upper bound on the number of experiments the proposed algorithm needs to carry out before identifying the optimal arm. Before stating the result, let us take $\Delta_{\min} = \min_{i\in[K]}\Delta_{i}$ to be the smallest gap and for any $\epsilon>0$,  define  
\begin{equation}
\label{Hedef}
H_\epsilon = \frac{18Kk_\mu}{\max\left(3\epsilon,\epsilon +  \Delta_{\min}\right)^2},
\end{equation}
which represents the complexity of the exploration problem in terms of the problem parameters. 
\begin{theorem}
\label{compthm}
Let $\tau$ be the stopping time of the proposed algorithm. Then 
\begin{equation}
\label{sample1}
\tau \leq \frac{64d\kappa^2R^2}{c_\mu^2}H_\epsilon\left(\log\left[\frac{64d\kappa^2R^2}{c_\mu^2}H_\epsilon\left(\pi\sqrt{\frac{d}{6\delta}}+1\right)\right] +\frac{c_\mu}{4\kappa R}\sqrt{\frac{K+1}{dH_\epsilon}}\right)^2
\end{equation}
is satisfied with probability at least $1-\delta$. In asymptotic notation, \eqref{sample1} can be expressed as 
\begin{equation}
\label{sample2}
\tau = O\left(\frac{d\kappa^2R^2}{c_\mu^2}H_\epsilon\left[\log\left(\frac{d^{3/4}\kappa R H_\epsilon^{1/2}}{c_\mu\delta^{1/4}}\right)\right]^2\right).
\end{equation}
\end{theorem}

Theorem \ref{compthm} provides an upper bound on the stopping time of the proposed algorithm in terms of the parameters of the exploration problem. As expected,  the number of experiments required by the proposed algorithm before declaring a near-optimal arm decreases in the reward tolerance ($\epsilon$) and the error probability ($\delta$), and increases in the number of features ($d$) and arms ($K$). In terms of the dependence on dimension ($d$), reward bound ($R$) and the complexity parameter ($H_\epsilon$), the sample complexity in \eqref{sample2} is  similar to that derived in \cite{xu2018} for linear bandits. The main difference is the appearance of the factor $\kappa/c_\mu$ in the complexity bound \eqref{sample2}, which encodes the difficulty of learning the inverse link function $\mu$. As the inverse link function $\mu$ becomes flatter on the boundaries of the input domain (i.e., has smaller $c_\mu$), more samples are required to distinguish between different pairs of arms.

The remainder of this subsection is devoted to proving Theorem \ref{compthm}, which  requires a set of preliminaries. We start by introducing some additional notation. For any $i,j\in[K]$ and any $t$, let 
$$(c_1^{ij}(t),c_2^{ij}(t))=\argmin_{c,c'\in[c_\mu,k_\mu]}{~\|cx^i - c'x^j\|_{M_t^{-1}}},$$
and define $y^{ij}_t = c_1^{ij}(t)x^i - c_2^{ij}(t)x^j$. Note that with this notation,  $y_t$ defined in Algorithm \ref{alg2} can be represented as $y_t = c_1^{i_tj_t}(t)x^{i_t} - c_2^{i_tj_t}(t)x^{j_t}$. For any $y\in\R^d$, let $\rho(y)$ be the optimal value of the linear program in \eqref{linprog}; i.e.,
\begin{equation}
\label{eqapp1}
\rho(y) = \sum_{a=1}^K |w^*_a(y)|.
\end{equation}
For any two real numbers $a,b$, we use the shorthand notation $a\vee b = \max(a,b)$. 

Our proof for Theorem \ref{compthm} relies on a number of results from the literature, which we state here for completeness. The following two lemmas are proved in \cite{xu2018}.
\begin{lemma}[Lemma 1 of \cite{xu2018}]
\label{lem1}
For any $y\in\R^d$, we have
$$\|y\|_{M_t^{-1}}\leq \sqrt{\frac{\rho(y)}{T_{y}(t)}},$$
where 
$$T_{y}(t) = \min_{k\in[K]:~p_k(y)>0}~\frac{T_k(t)}{p_k(y)}.$$
\end{lemma}

\begin{lemma}[Lemma 4 of \cite{xu2018}]
\label{lem2}
When event $\EE$ holds, $B(t)$ satisfies the following bounds:
\begin{enumerate}
\item If either $i_t$ or $j_t$ is the best arm:
$$B(t)\leq \min(0,-(\Delta_{i_t}\vee \Delta_{j_t}) + \beta_t(i_t,j_t)) + \beta_t(i_t,j_t),$$
\item If neither $i_t$ nor $j_t$ is the best arm:
$$B(t)\leq \min(0,-(\Delta_{i_t}\vee \Delta_{j_t}) + 2\beta_t(i_t,j_t)) + \beta_t(i_t,j_t).$$
\end{enumerate}
\end{lemma}
The following lemma is proved in \cite{antos2010}.
\begin{lemma}[Proposition 6 of \cite{antos2010}]
\label{lem4}
For any $t>0$, let $q(t) = a\sqrt{t} + b$ and $l(t) = \log(t)$, for some $a>0$. Define $t^* = \frac{4}{a^2}\left[\log\Bigl(\frac{4}{a^2}\Bigr) - b\right]^2$. Then, for any positive $t$ such that $t\geq t^*$, we have $q(t)>l(t)$. 
\end{lemma}

The following lemma establishes an upper bound on the solution of the linear program in \eqref{linprog}.
\begin{lemma}
\label{lem3}
Let $y_t = c_1^{i_tj_t}(t)x^{i_t} - c_2^{i_tj_t}(t)x^{j_t}$ and let $w^*(y_t)$ be the solution of \eqref{linprog}. Then we have
$$\|w^*(y_t)\|_\infty \leq 2k_\mu.$$
\end{lemma}
\proof{Proof.}
Let $w'\in\R^K$ be such that $w'_{i_t} = c_1^{i_tj_t}(t), w'_{j_t} = -c_2^{i_tj_t}(t)$ and all other elements of $w'$ are zero. Clearly, $w'$ satisfies the constraint in \eqref{linprog}. Therefore, we have
$$\|w^*(y_t)\|_\infty\leq\|w^*(y_t)\|_1 \leq \|w'\|_1 = |c_1^{i_tj_t}(t)| + |c_2^{i_tj_t}(t)|\leq k_\mu +k_\mu = 2k_\mu.$$ \Halmos
\endproof
With the above lemmas in place, we are ready to prove the following theorem.
\begin{theorem}
\label{thm4}
If event $\EE$ occurs, then the stopping time of the proposed algorithm satisfies 
$$\tau\leq H_\epsilon C_\tau^2 + K +1.$$
\end{theorem}

\proof{Proof.}
Suppose that event $\EE$ occurs. Let $k\in[K]$ be an arbitrary arm, and let $t_k<\tau$ be the last round in which arm $k$ was pulled. Because $B(t_k)>\epsilon$, Lemma \ref{lem2} implies that 
$$\min\left(0, -(\Delta_{i_{t_k}}\vee\Delta_{j_{t_k}})+2\beta_{t_k}(i_{t_k},j_{t_k})\right) + \beta_{t_k}(i_{t_k},j_{t_k})\geq B(t_{k})\geq \epsilon,$$
which in turn gives rise to the following three inequalities:
\begin{equation}
\label{eq21a}
\epsilon\leq \beta_{t_k}(i_{t_k},j_{t_k}),~~ \epsilon + \Delta_{i_{t_k}}\leq 3\beta_{t_k}(i_{t_k},j_{t_k}),~~\epsilon + \Delta_{j_{t_k}}\leq 3\beta_{t_k}(i_{t_k},j_{t_k}).
\end{equation} 
Rearranging~(\ref{eq21a}) yields 
\begin{equation}
\label{betabound}
\beta_{t_k}(i_{t_k},j_{t_k})\geq \max\left(\epsilon,\frac{\epsilon +  \Delta_{i_{t_k}}}{3},\frac{\epsilon +  \Delta_{j_{t_k}}}{3}\right).
\end{equation}

From Lemma \ref{lem1}, we have
$$T_{y_{t_k}}({t_k})\leq \frac{\rho(y_{t_k})}{\|y_{t_k}\|^2_{M_{t_k}^{-1}}},$$
which by substituting $\beta_{t_k}(i_{t_k},j_{t_k}) = C_{t_k} \|y_{t_k}\|_{M_{t_k}^{-1}}$ from~(\ref{betadef}) gives
\begin{equation}
\label{Tbound}
T_{y_{t_k}}(t_k)\leq \frac{\rho(y_{t_k}) C_{t_k}^2}{\beta_{t_k}(i_{t_k},j_{t_k}) ^2}.
\end{equation}
Combining \eqref{betabound} and \eqref{Tbound} gives
\begin{equation}
\label{Tbound2}
T_{y_{t_k}}({t_k})\leq \frac{\rho(y_{t_k}) C_{t_k}^2}{\max\left(\epsilon,\frac{\epsilon +  \Delta_{i_{t_k}}}{3},\frac{\epsilon +  \Delta_{j_{t_k}}}{3}\right)^2}.
\end{equation} 

Because arm $k$ was selected by Algorithm \ref{alg3} in round $t_k$, it follows that 
\begin{equation}
\label{eq24a}
T_k(t_k) = p_k(y_{t_k})T_{y_{t_k}}(t_k).
\end{equation}
Then, from \eqref{Tbound2}, we get 
\begin{align}
T_k(\tau) &= T_k(t_k) + 1 &&{\rm by~construction},\nonumber\\
& = p_k(y_{t_k})T_{y_{t_k}}(t_k)  +1 &&{\rm by} ~(\ref{eq24a}),\nonumber\\
& \leq \frac{p_k(y_{t_k})\rho(y_{t_k}) }{\max\left(\epsilon,\frac{\epsilon +  \Delta_{i_{t_k}}}{3},\frac{\epsilon +  \Delta_{j_{t_k}}}{3}\right)^2}C_{t_k}^2 + 1 &&{\rm by} ~(\ref{Tbound2}),\nonumber\\
& = \frac{|w^*_k(y_{t_k})| }{\max\left(\epsilon,\frac{\epsilon +  \Delta_{i_{t_k}}}{3},\frac{\epsilon +  \Delta_{j_{t_k}}}{3}\right)^2}C_{t_k}^2 +1 &&{\rm by} ~(\ref{prel})  ~{\rm and} ~(\ref{eqapp1}),\nonumber\\
& \leq \frac{\|w^*(y_{t_k})\|_\infty }{\max\left(\epsilon,\frac{\epsilon +  \Delta_{i_{t_k}}}{3},\frac{\epsilon +  \Delta_{j_{t_k}}}{3}\right)^2}C_{t_k}^2+1,&& \nonumber\\
& \leq \frac{2k_\mu}{\max\left(\epsilon,\frac{\epsilon +  \Delta_{i_{t_k}}}{3},\frac{\epsilon +  \Delta_{j_{t_k}}}{3}\right)^2}C_{t_k}^2+1 &&{\rm by ~Lemma} ~\ref{lem3}, \nonumber\\
& \leq \frac{2k_\mu}{\max\left(\epsilon,\frac{\epsilon +  \min_{i\in[K]}\Delta_{i}}{3}\right)^2}C_{t_k}^2+1 && \nonumber\\
&  \leq \frac{18k_\mu}{\max\left(3\epsilon,\epsilon + \Delta_{\mbox{min}}\right)^2}C_{t_k}^2+1 &&\nonumber\\
&\leq \frac{18k_\mu}{\max\left(3\epsilon,\epsilon + \Delta_{\mbox{min}}\right)^2}C_{\tau}^2+1. && \label{Tbound3}
\end{align}

Note that $\tau = 1 + \sum_{k=1}^K T_k(\tau)$. Hence, it follows from~(\ref{Hedef}) and~\eqref{Tbound3} that 
$$\tau\leq H_\epsilon C_\tau^2 + K +1,$$
which completes the proof. \Halmos
\endproof

We are now in a position to prove Theorem \ref{compthm}. 
\proof{Proof of Theorem \ref{compthm}.}
Suppose that event $\EE$ holds. Then
\begin{align}
\tau &\leq H_\epsilon C_\tau^2 +  K +1 ~~{\rm by ~Theorem} ~\ref{thm4},\nonumber\\
&= \frac{16d\kappa^2R^2}{c_\mu^2}H_\epsilon \log(\tau)\log\left(\pi\sqrt{\frac{d}{6\delta}}\tau\right) + K+1 ~~{\rm by} ~(\ref{confwidth}),\nonumber\\
&\le  \frac{16d\kappa^2R^2}{c_\mu^2}H_\epsilon \left(\log\left[\left(\pi\sqrt{\frac{d}{6\delta}}+1\right)\tau\right]\right)^2 + K+1,\label{ineq1}
\end{align} 
where~(\ref{ineq1}) follows by Jensen's inequality and the logarithmic arithmetic-geometric mean inequality (i.e., $\sqrt{\log x\log y}\le \frac{1}{2}(\log x  + \log y)\le \log\Bigl(\frac{x+y}{2}\Bigr)$). 
Applying the inequality $x+y\leq (\sqrt{x}+\sqrt{y})^2$ for any $x,y>0$ to \eqref{ineq1} yields 
\begin{equation}
\label{ineq2}
\sqrt{\tau}\leq \frac{4\kappa R\sqrt{d}}{c_\mu}\sqrt{H_\epsilon}\log\left[\left(\pi\sqrt{\frac{d}{6\delta}}+1\right)\tau\right] + \sqrt{K+1}.
\end{equation}

Let $c_1 = \frac{4\kappa R\sqrt{d}}{c_\mu}\sqrt{H_\epsilon}, ~c_2 =\pi\sqrt{\frac{d}{6\delta}}+1$ and define $a = \frac{1}{c_1\sqrt{c_2}}, ~b = -\frac{\sqrt{K+1}}{c_1}$.  With a change of variable $z = c_2\tau$, \eqref{ineq2}  can be written as 
$$a\sqrt{z} + b \leq \log(z).$$
According to Lemma \ref{lem4}, this is possible only if $$z\leq z^* = \frac{4}{a^2}\left[\log\left(\frac{4}{a^2}\right) - b\right]^2.$$
Substituting for $a,b$ and $z$ shows that \eqref{ineq2} holds only if 
$$\tau\leq \frac{64d\kappa^2R^2}{c_\mu^2}H_\epsilon\left(\log\left[\frac{64d\kappa^2R^2}{c_\mu^2}H_\epsilon\left(\pi\sqrt{\frac{d}{6\delta}}+1\right)\right] +\frac{c_\mu}{4\kappa R}\sqrt{\frac{K+1}{dH_\epsilon}}\right)^2.$$
On the other hand, according to Theorem \ref{prop1}, event $\EE$ holds with probability at least $1-\delta$. This completes the proof.\Halmos
\endproof

\section{Numerical Results}
\label{sec::sim}
In this section, we test the proposed algorithm in various scenarios using synthetic data in Subsection~\ref{ssec-synthetic} and drug design data in Subsection~\ref{ssec-real}. The experimental details are described in Subsection~\ref{ssec-parameters}. 

\subsection{Experimental Details}\label{ssec-parameters}
All scenarios considered here involve a logistic inverse link function with binary observations. Recall that the proposed algorithm starts with an exploratory phase in which $E$ random arms are played to ensure that $M_{E+1}$ is nonsingular.  Although we have not optimized this parameter in our numerical studies, we use $E=\min(K,3d)$ throughout all simulations here.

The confidence sets defined in \eqref{confwidth} and \eqref{gapconfdef} rely on the free parameter $\alpha$. Our theoretical analysis in Section \ref{sec::theo} suggests the use of $\alpha=\frac{2\kappa R}{c_{\mu}}$ from~(\ref{alpha}). However, the theoretical confidence sets derived for generalized linear settings in \cite{filippi2010} are loose and the performance can be  improved by scaling these sets. Although Section 4.2 of \cite{filippi2010} uses an asymptotic analysis to shrink these sets in the regret-minimization setting (a finite-sample result in Theorem 1 of \cite{li2017} provides a rigorous justification of this analysis), we simply note that all the rewards and reward gaps are in $[0,1]$ in the case of a logistic reward function while the  width of the  confidence set defined in \eqref{gapconfdef} may be larger than one. Consequently, we set $\alpha$ so as to constrain the confidence set widths in \eqref{gapconfdef} to the interval $[0,1]$, which can be achieved by setting $\frac{2\kappa R}{c_{\mu}}$ times $\max_{i,j\in [K]}\beta_{E(i,j)}$ equal to 1, yielding via \eqref{confwidth} and \eqref{betadef}
\begin{equation}
\label{eqalpha0}
\alpha = \frac{1}{ \frac{2\kappa R}{c_\mu}\sqrt{2d\log E\log\Bigl(\frac{\pi^2dE^2}{6\delta}\Bigr)}\max_{i,j\in[K]}\max_{c,c'\in[c_\mu,k_\mu]} \|cx^i-c'x^j\|_{M_E^{-1}}},
\end{equation}
which is determined at the end of the exploratory phase of length $E$. 
 
Throughout our simulations, we fix the confidence parameter $\delta = 0.05$, which theoretically guarantees the accuracy of the proposed algorithm with $95\%$ confidence.  The results in Figs.~\ref{fig0}-5 are generated by averaging over tens of random realizations of the problem. In all cases, the proposed algorithm achieved the desired accuracy (as dictated by $\epsilon$) with probability more than 0.95.

\subsection{Synthetic Data Simulations}\label{ssec-synthetic}
To create a synthetic data set, we generate the reward parameter $\theta$ by sampling from a $d-$dimensional standard multivariate Gaussian distribution. The feature vector associated with each of the $K$ arms is randomly generated such that each of its components is uniformly distributed in $[-1,1]$. Acting as if we do not know the reward parameter $\theta$, we run the proposed algorithm to identify the best among the available $K$ arms under different scenarios. Recalling that the parameters $L$, $R$ and $S$ are defined via $\|x\|_2\leq L$, $r_t\le R$ and $\|\theta\|_2\leq S$, we have $L\le d$ and $R=1$. Note that the parameter $S$ is used only in determining $c_{\mu}$, which we treat as given in our simulations; hence, there is no need to specify $S$.

\begin{figure}[t]
\begin{center}
\includegraphics[height=2in]{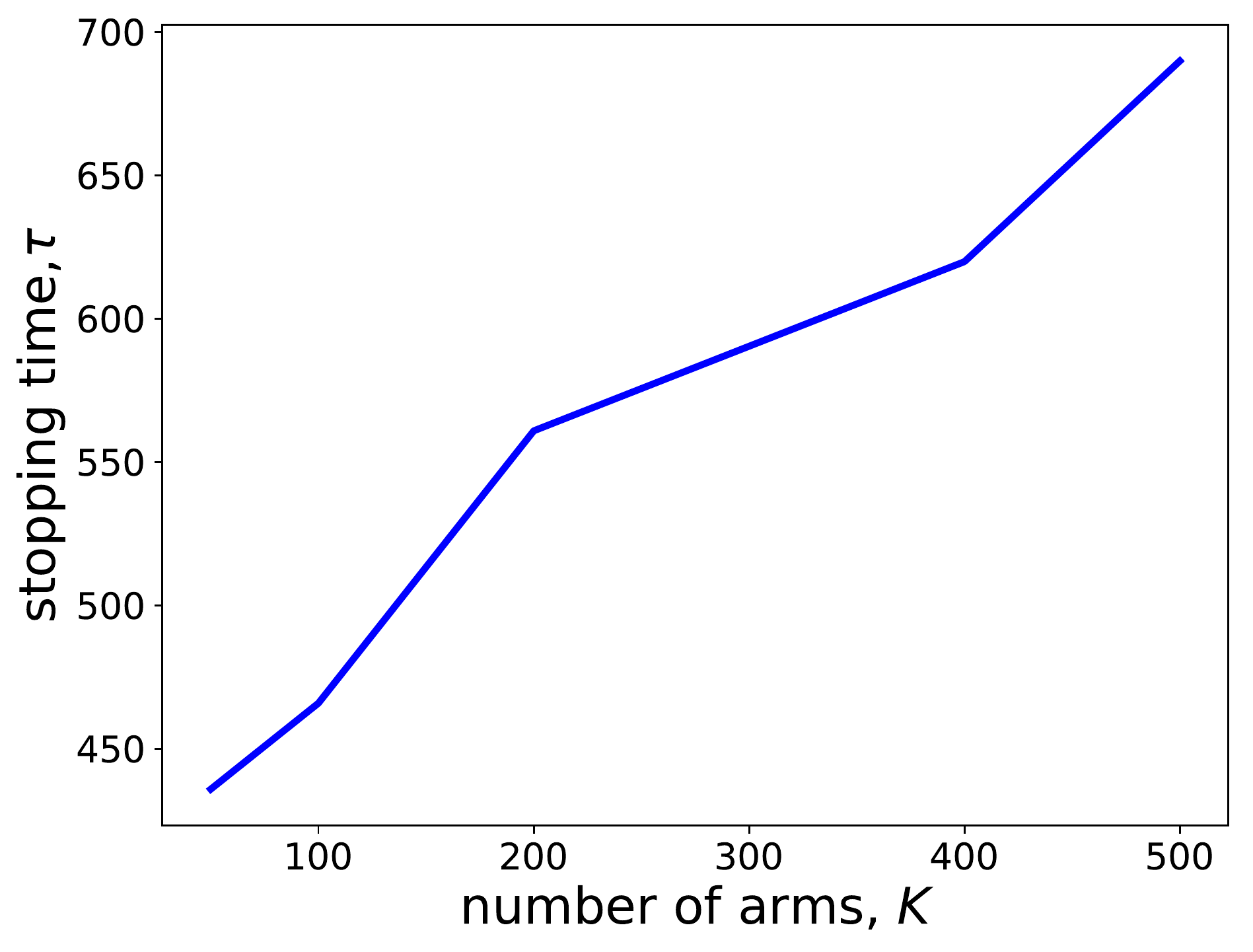}
\caption{The amount of exploration required by the proposed algorithm versus the number of available arms, with tolerance $\epsilon=0.1$, confidence parameter $\delta=0.05$ and  $d=10$ features.} \label{fig0}
\end{center}
\end{figure}

\begin{figure}[t]
\begin{center}
\includegraphics[height=2in]{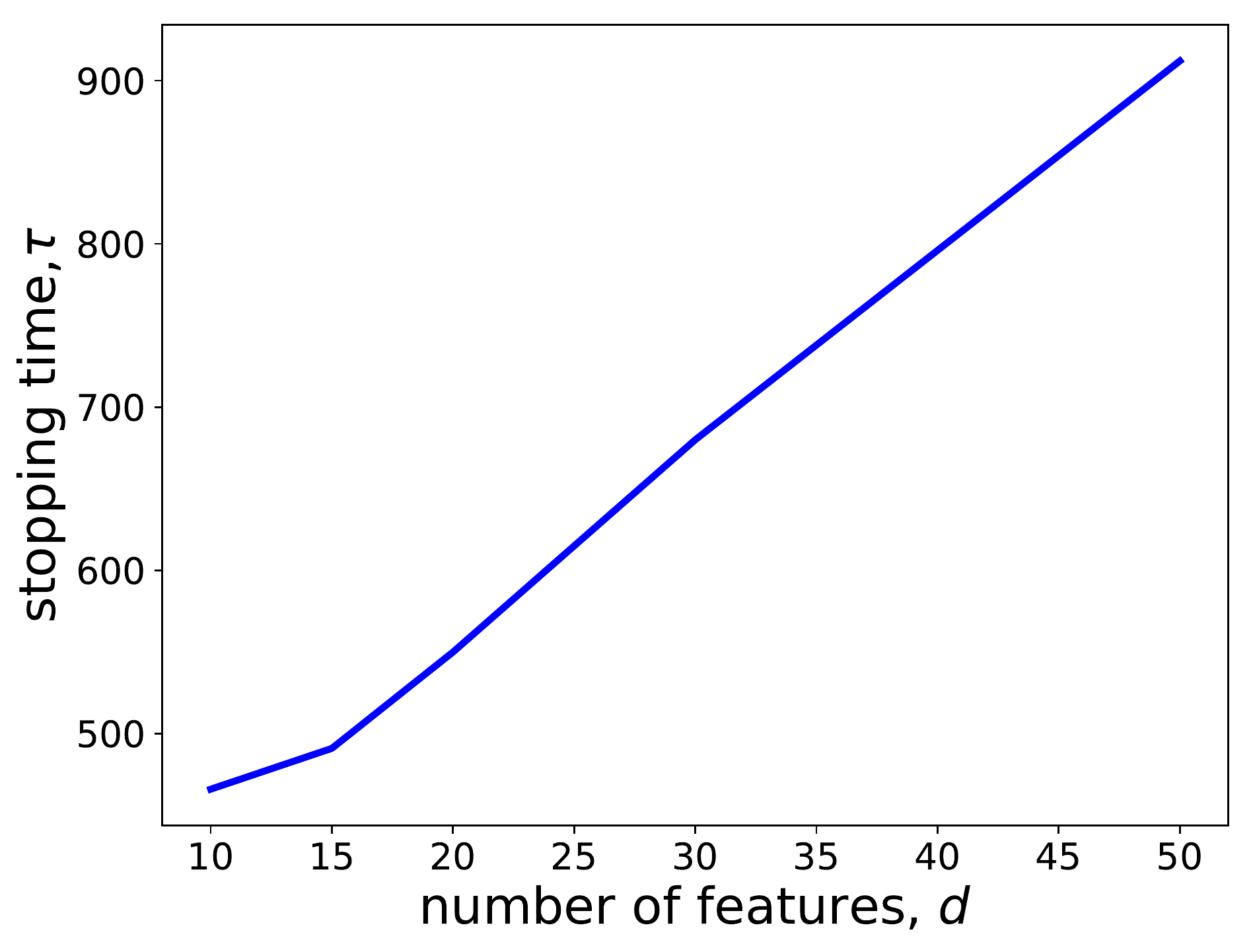}
\caption{The amount of exploration required by the proposed algorithm versus the number of features, with tolerance $\epsilon=0.1$, confidence parameter $\delta=0.05$ and  $K=10$ arms.} \label{fig::syn_dim}
\end{center}
\end{figure}

\begin{figure}[t]
\begin{center}
\includegraphics[height=2in]{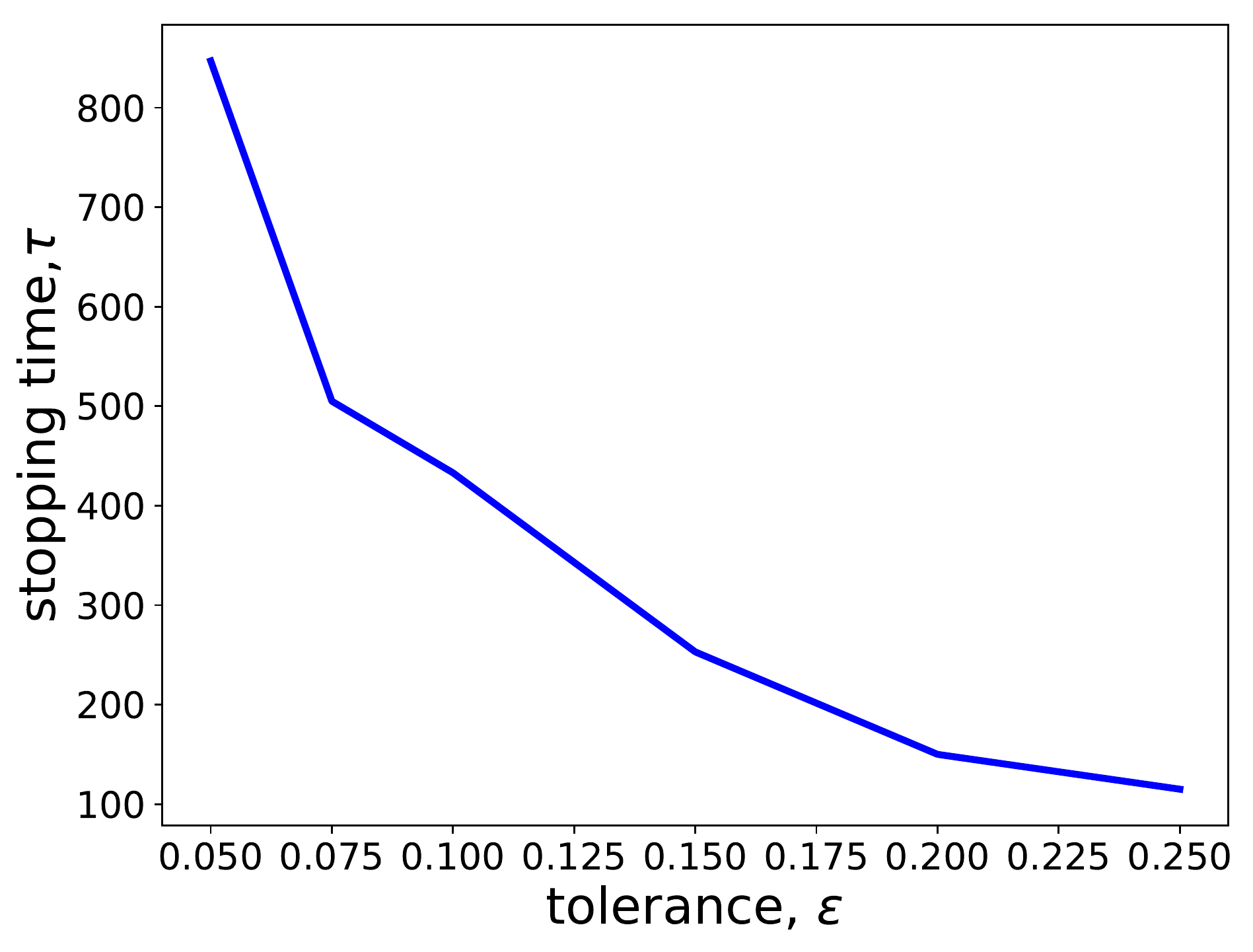}
\caption{The amount of exploration required by the proposed algorithm versus the tolerance level $\epsilon$, with confidence parameter $\delta=0.05$ $K=100$ arms, and $d=10$ features.} \label{fig1}
\end{center}
\end{figure}

We first fix the tolerance parameter $\epsilon=0.1$ and the feature dimension $d=10$, and plot the stopping time $\tau$ as a function of the number of arms $K$ (Fig.~\ref{fig0}). We then fix $\epsilon=0.1$ and the number of arms $K=100$, and compute $\tau$ as a function of the feature dimension $d$ (Fig.~\ref{fig::syn_dim}). These plots reveal that the amount of exploration undertaken by the proposed algorithm depends more strongly on the feature dimension $d$ than on the number of arms $K$. This is due to the fact that the proposed algorithm builds confidence sets for the reward parameter, which is of dimension $d$, rather than for the reward of each arm separately. 

Fixing the feature dimension $d=10$ and the number of arms $K=100$, we plot the stopping time $\tau$ versus the tolerance parameter $\epsilon$ (Fig.~\ref{fig1}). As expected, a smaller tolerance requires a larger amount of exploration in order to guarantee that the identified arm is within the acceptable distance of the optimal arm. 

Recall that exploration algorithms for independent arms require playing each arm many times. To assess the improvement over this approach, we implement the GapE algorithm  \citep{gabillon2012}, which treats arms as independent with binary rewards, to an instance of this synthetic scenario with $K=50$ arms, $d=10$ features and $\epsilon=0.1$. On average, the GapE algorithm requires 60k experiments before identifying a near-optimal arm, while our proposed algorithm identifies a near-optimal arm  after only 436 experiments. 

We know of no other algorithms that are natural comparisons to our proposed algorithm: the algorithms in \cite{xu2018} and \cite{soare2014} use a linear model that does not accommodate binary outcomes, and regret-minimization algorithms are not typically tested against best-arm identification algorithms because the former do not have any stopping criteria, which precludes us from comparing our algorithm to the one in \cite{filippi2010}.  

\subsection{Drug Design Simulations}\label{ssec-real}
In this subsection, we consider the problem of identifying the most effective among a large set of designed drugs for a certain disease. We use the ChEMBL data repository \citep{gaulton2011} and download the data set describing 982 small molecules designed for targeting the Cyclin-Dependent Kinase 2/Cyclin A protein complex. The covariates for each small molecule are associated with its three-dimensional structure, which can be represented as a 1024-dimensional binary vector. To construct such a representation, we compute the Morgan Fingerprint \citep{morgan} of each molecule with radius 4. Then, by applying a Principal Components Analysis (PCA) to these sparse high-dimensional binary vectors, we generate a feature vector with the desired dimension. 

The generalized linear model allows for the random experimental output to have values that are continuous, integer or binary, depending on the link function that is used. If we want to use a model with binary data (i.e., the logit inverse link function), we would need binomial or Bernoulli data for each arm. However, we have been unable to find data like these. Rather, the output data in the data set from \cite{gaulton2011} are the affinity of each small molecule, as measured by the IC$_{50}$ (the concentration of an inhibitor where the biological function, typically the binding to a protein, is reduced by half). The IC$_{50}$ is a deterministic number (there are no data in \cite{gaulton2011} on the possible measurement errors) that is an indicator of the effectiveness of each small molecule. 

To adapt these outcome data into our modeling framework, we convert these deterministic IC$_{50}$ values into random binary outcomes as follows. First, we assume the drug produces a cure if the IC$_{50}$ value is less than $10^6$nM (nanomolar), and does not produce a cure if the IC$_{50}$ is greater than or equal to $10^6$nM. Next, given the feature vectors and these artificial cure outcomes, we fit a logistic regression model and estimate the cure rate of each drug. Finally, by assuming that these estimated cure rates are the unknown expected rewards of the drugs, we apply the proposed algorithm to find the best drug; i.e., in our  simulations, each time a drug is tested, we observe  a Bernoulli random variable with its mean equal to the drug's expected reward. 

\begin{figure}[t]
\begin{center}
\includegraphics[height=2in]{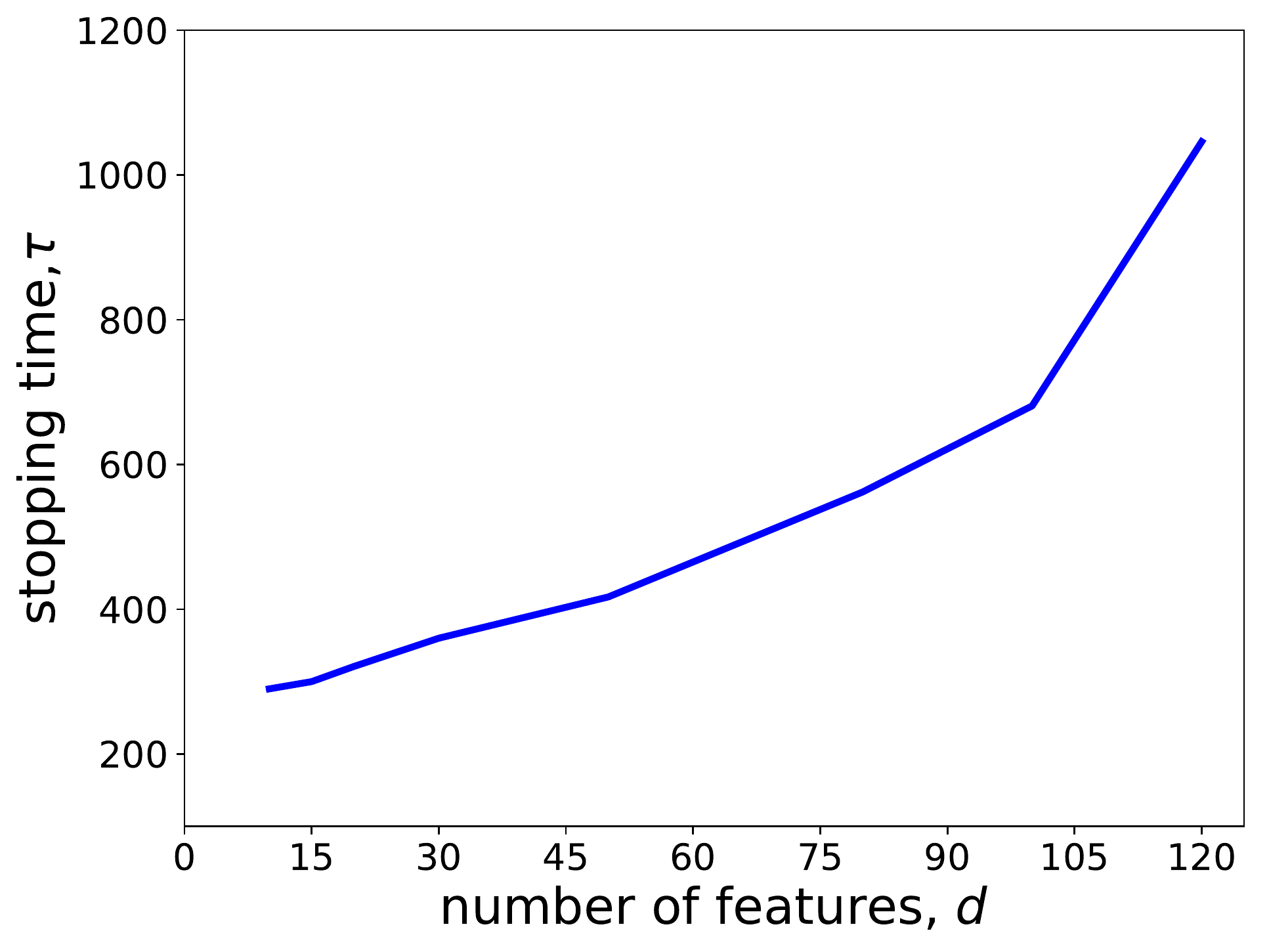}
\caption{The amount of exploration required by the proposed algorithm versus the number of features in the drug design simulations, with tolerance $\epsilon=0.1$, confidence parameter $\delta=0.05$ and $K=400$ arms.} \label{fig2}
\end{center}
\end{figure}

We first fix the tolerance parameter $\epsilon=0.1$ and the number of arms $K=400$ by randomly choosing 400 of the 982 molecules in \cite{gaulton2011}. The stopping time $\tau$ is plotted against the feature dimension $d$, which is varied using PCA (Fig.~\ref{fig2}). As in the synthetic data simulations, this relationship is increasing and -- when the number of features is smaller than 50 -- the proposed algorithm finishes exploration before all the arms are explored.

To roughly assess the effectiveness of the proposed algorithm in its choice of best arm, we note that the 982 IC$_{50}$ values in the data set range from 0.3nM to $1.5\times 10^6$nM, and that the best (i.e., lowest-valued) among the 400 molecules in the experiment has an IC$_{50}=0.4$nM. The algorithm's declared optimal arm has an IC$_{50}=11.99$nM. However, because considerable information is lost when we binarize the data (e.g., two drugs with IC$_{50}$ values of 0.5nM and $10^5$nM would both be assigned a binary value of one for a cure), we should not expect our algorithm to identify the drug with the lowest IC$_{50}$. 

\begin{figure}[t]
\begin{center}
\includegraphics[height=2in]{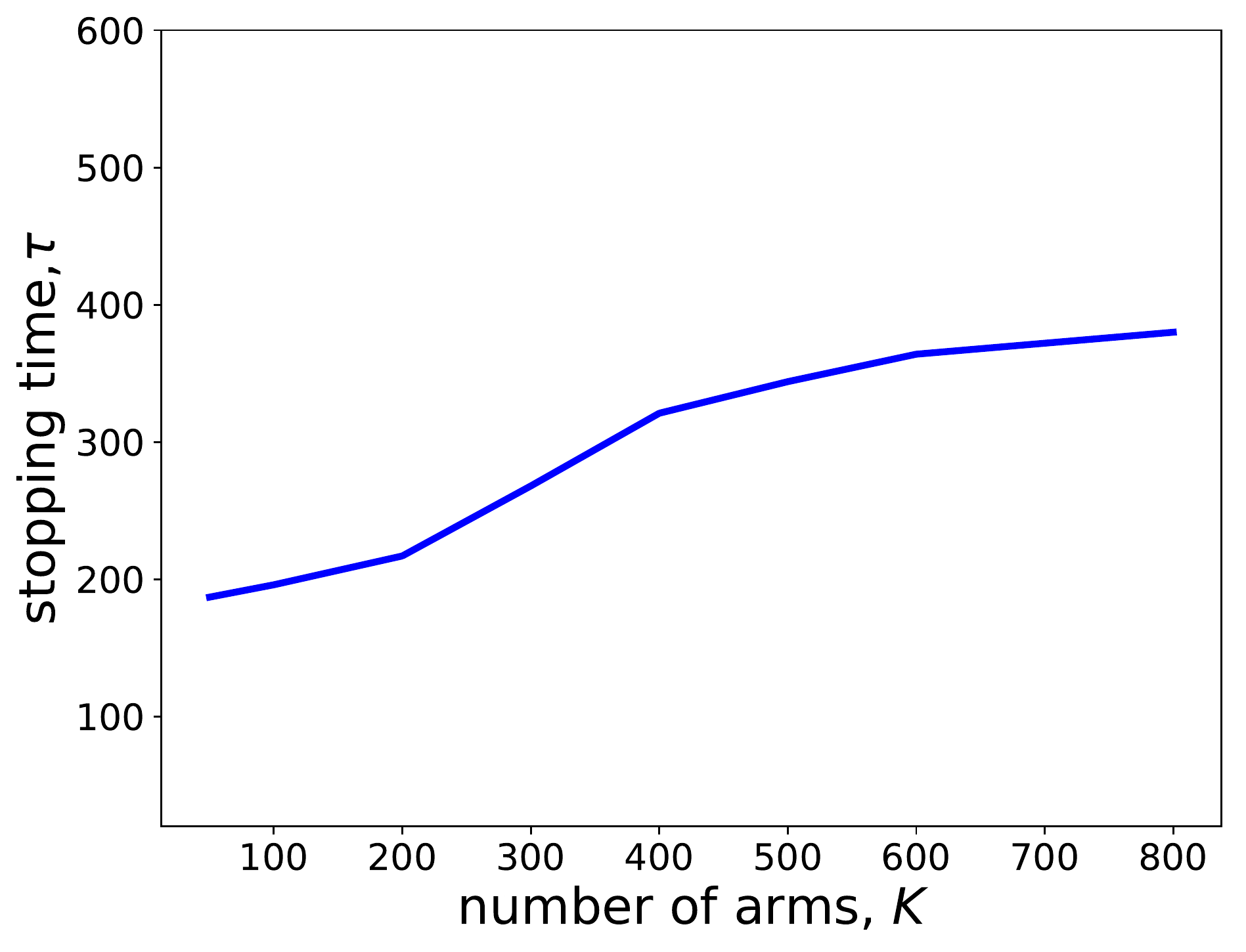}
\caption{The amount of exploration required by the proposed algorithm versus the number of arms in the drug design simulations, with tolerance $\epsilon=0.1$, confidence parameter $\delta=0.05$ and $d=20$ features.} \label{fig3}
\end{center}
\end{figure}

We next fix $\epsilon=0.1$ and the feature dimension $d=20$, and plot the stopping time $\tau$ versus the number of arms $K$ (Fig.~\ref{fig3}), where for each value of $K$ we randomly choose $K$ arms out of the 982 small molecules in \cite{gaulton2011}.  When there are more than 200 arms in Fig.~\ref{fig3}, the number of experiments carried out by the proposed algorithm is smaller than the total number of arms. 

\section{Conclusion}
\label{sec::concl}
Motivated by drug design, we consider the best-arm identification problem for a generalized linear bandit, where arms correspond to drugs, the covariate vectors describe biological properties of the drugs, and the inverse link function in the generalized linear model captures the nonlinear and perhaps binary relationship between the covariates and the experimental outcomes. We use ideas from \cite{filippi2010} on building confidence sets, but apply these directly to confidence sets for gaps between pairs of arms rather than confidence sets for rewards of arms, which allows us to generalize the results in \cite{xu2018} from the linear bandit model to the generalized linear bandit model. After performing our analysis, we became aware of recent work \citep{li2017} that improves the regret rate in \cite{filippi2010} by a factor of $\sqrt{d}$. We leave it to future work to investigate whether the results in \cite{li2017} can be used to improve our theoretical bounds. 

Numerical results show that -- at least with $\delta=0.05$ and $\epsilon = 0.1$ (i.e., being 95\% confident of selecting an arm with a cure rate within 10\% of optimal) -- can be identified very quickly: when there are a large number of arms, the number of experiments performed can be much less than the total number of arms (Fig.~\ref{fig3}). This is achieved by learning about all arms whenever an arm is played. Our algorithm, coupled with an approach to dimensionality reduction of the feature dimension such as PCA or machine learning tools, has the potential to streamline some nonlinear problems in drug design and other experimental settings. 
%
%

\ACKNOWLEDGMENT{This research was supported by the Graduate School of Business, Stanford University (L.M.W.) and a Stanford Graduate Fellowship (A.K.).}





\end{document}